%% file: emnlp2022.tex
\pdfoutput=1

\documentclass[11pt]{article}

\usepackage[]{EMNLP2022}

\usepackage{times}
\usepackage{latexsym}

\usepackage[T1]{fontenc} 
\usepackage{CJKutf8} 

\usepackage[utf8]{inputenc}
\usepackage[polutonikogreek,english]{babel}

\usepackage{microtype}

\usepackage{url}
\usepackage{multirow}
\usepackage{booktabs}
\usepackage{tabularx}
\usepackage{graphics}
\usepackage{graphicx}
\usepackage{colortbl}
\usepackage{xcolor,soul}
\usepackage{makecell}
\usepackage{subcaption}

\definecolor{berry}{HTML}{BC3754}
\definecolor{berry2}{HTML}{e897a8}
\definecolor{grey}{HTML}{D3D3D3}

\sethlcolor{berry2}

\title{Language Contamination Helps Explain the Cross-lingual\\ Capabilities of English Pretrained Models}

\author{Terra Blevins$^{1}$ \quad Luke Zettlemoyer$^{1,2}$ \\
        $^{1}$ Paul G. Allen School of Computer Science \& Engineering, University of Washington \\
        $^{2}$ Meta AI Research \\
        {\tt \{blvns, lsz\}@cs.washington.edu}}

\begin{document}
\maketitle
\begin{abstract}
English pretrained language models, which make up the backbone of many modern NLP systems, require huge amounts of unlabeled training data.
These models are generally presented as being trained only on English text but have been found to transfer surprisingly well to other languages.
We investigate this phenomenon and find that common English pretraining corpora actually contain significant amounts of non-English text: even when less than 1\% of data is not English (well within the error rate of strong language classifiers), this leads to hundreds of millions of foreign language tokens in large-scale datasets.
We then demonstrate that even these small percentages of non-English data facilitate cross-lingual transfer for models trained on them, with target language performance strongly correlated to the amount of in-language data seen during pretraining.
In light of these findings, we argue that no model is truly monolingual when pretrained at scale, which should be considered when evaluating cross-lingual transfer.
\end{abstract}

\section{Introduction}
Pretrained language models have become an integral part of NLP systems. They come in two flavors: \textit{monolingual}, where the model is trained on text from a single language, and \textit{multilingual}, where the model is jointly trained on data from many different languages. Monolingual pretrained models are generally applied to tasks in the same language, whereas multilingual ones are used for cross-lingual tasks or transfer.

Recent work has claimed that monolingual pretrained models are also surprisingly good at transferring between languages, despite ostensibly having never seen the target language before \cite[inter alia]{gogoulou2021cross, li2021cross}.
However, because of the large scale of pretraining data and because many pretraining corpora are not publicly available, it is currently unknown how much foreign language data exists in monolingual pretraining corpora.
In this paper, we show that (1) these data are almost certainly contaminated with very small percentages of text from other languages and that (2) cross-lingual transfer is possible from such data leakage in the pretraining corpus.

\begin{figure}
    \centering
    \includegraphics[width=0.9\linewidth]{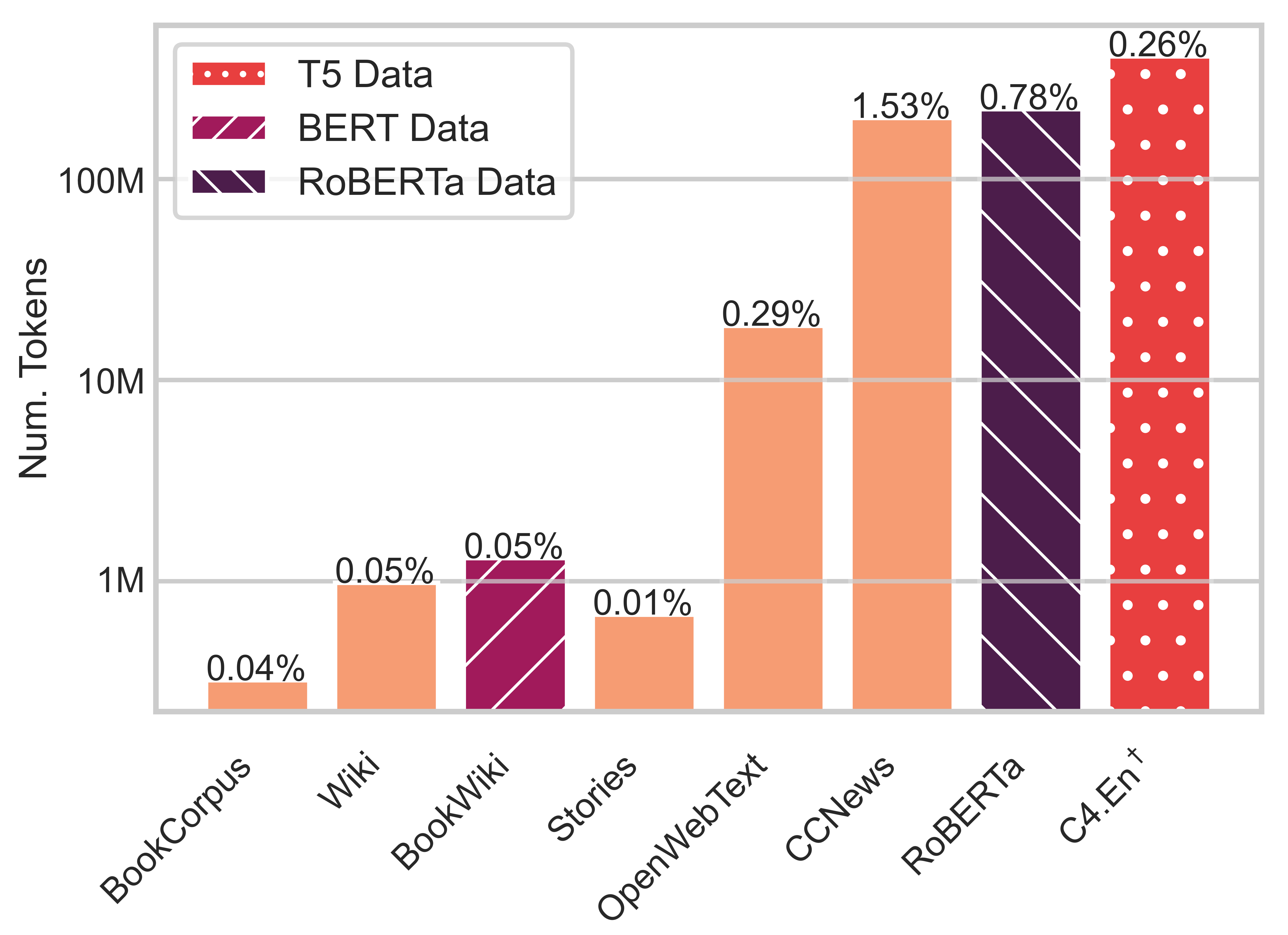}
    \vspace{-5pt}
    \caption{Estimated non-English data in English pretraining corpora (token count and total percentage); even small percentages lead to many tokens. C4.En ($\dagger$) is estimated from the first 50M examples in the corpus.
    }
    \label{fig:lang-comp-teaser}
\end{figure}

More specifically, we quantify how \textit{multilingual} English pretrained models are in two steps. First, we analyze common English pretraining corpora with a large-scale automatic evaluation to estimate their language composition, as well as a smaller-scale manual analysis. Second, we perform experiments across fifty languages on masked language modeling and part-of-speech (POS) tagging to measure how well the models trained on these pretraining corpora perform outside of English.

Our analysis finds that these corpora include very small percentages that amount to overall significant amounts of non-English text (Figure \ref{fig:lang-comp-teaser}), particularly those derived from web-crawled data. 
Furthermore, the models trained on this data perform surprisingly well on other languages; this transfer is strongly correlated with the amount of target language data seen during pretraining. Notably, we find that the English T5 outperforms mBERT on POS tagging in multiple languages with no finetuning. 

Overall, these results indicate that the considered models are actually multilingual and that their ability to transfer across languages is not zero-shot, despite what has been recently claimed.
Given the effort required to fully remove all non-English data, we question whether it is practically possible to train truly monolingual models at scale.

\section{Pretraining Data Composition}
\label{sec:lang-composition}

We first measure how much non-English text exists in commonly used English pretraining corpora with two analyses: an automatic language identification to estimate the amount of foreign language data in these corpora, and a manual qualitative analysis of the text classified as non-English.

We consider the following pretraining datasets: \textsc{English Wikipedia}
(11.8GB);  \textsc{BookCorpus} (\citealt{zhu2015aligning}, 4.2GB); \textsc{Stories} (\citealt{trinh2018simple}, 31GB); \textsc{OpenWebText} (\citealt{gokaslan2019OpenWeb}, 38GB), which is an open-source version of \textsc{WebText} \cite{radford2019language}; \textsc{CC-NEWS} (\citealt{liu2019roberta}, 76 GB); and \textsc{C4.En} (\citealt{raffel2020exploring}, 305GB), as provided by \citet{dodge2021documenting}. We use the versions of \textsc{Wikipedia}, \textsc{BookCorpus}, and \textsc{CC-NEWS} used to pretrain RoBERTa.

\begin{figure*}[t]
    \centering
    \begin{subfigure}{0.32\textwidth}
        \centering
        \includegraphics[width=\linewidth]{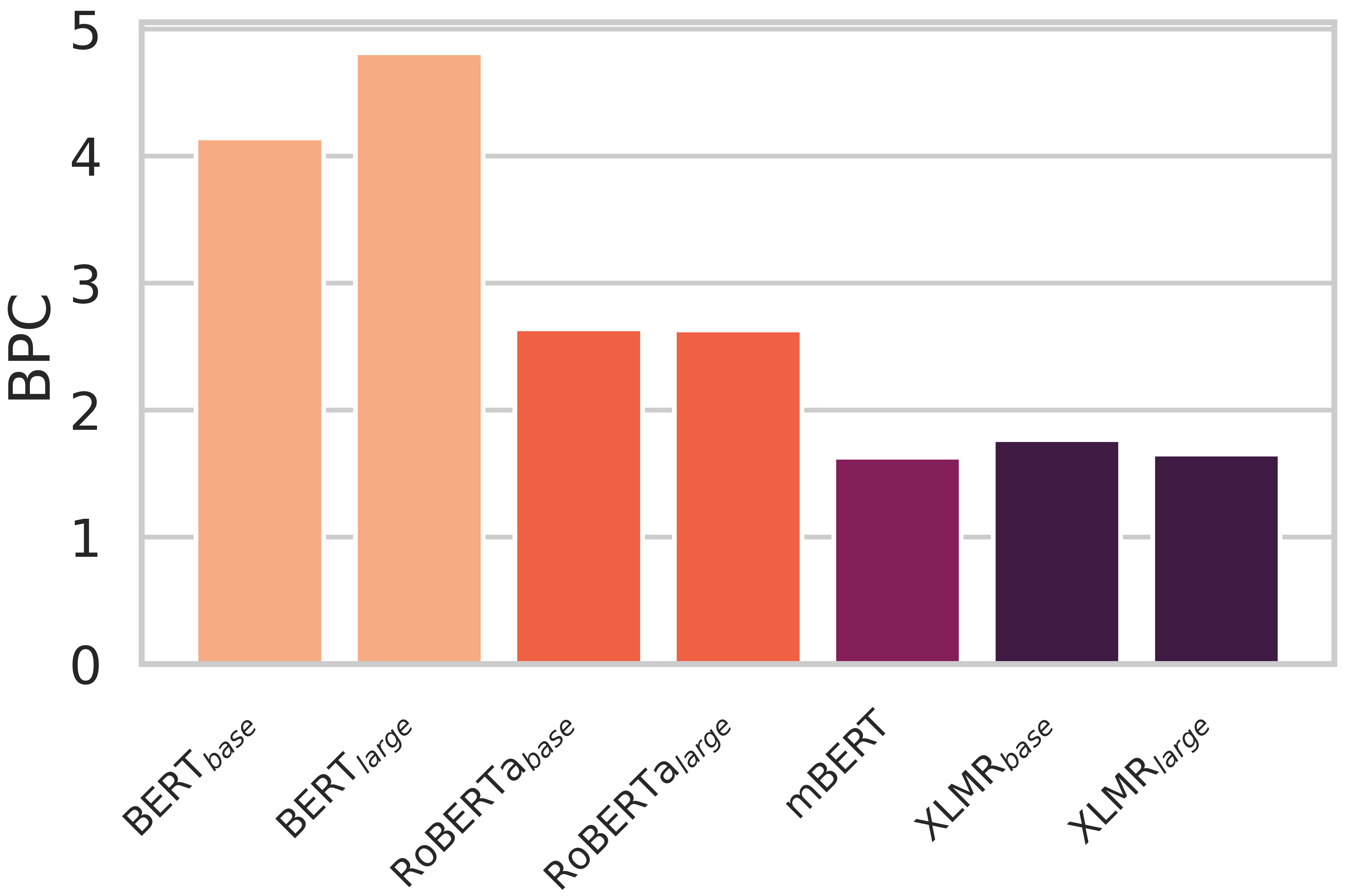} 
        \caption{MLM}
        \label{fig:summary-1}
    \end{subfigure}
    \hfill
    \begin{subfigure}{0.32\textwidth}
        \centering
        \includegraphics[width=\linewidth]{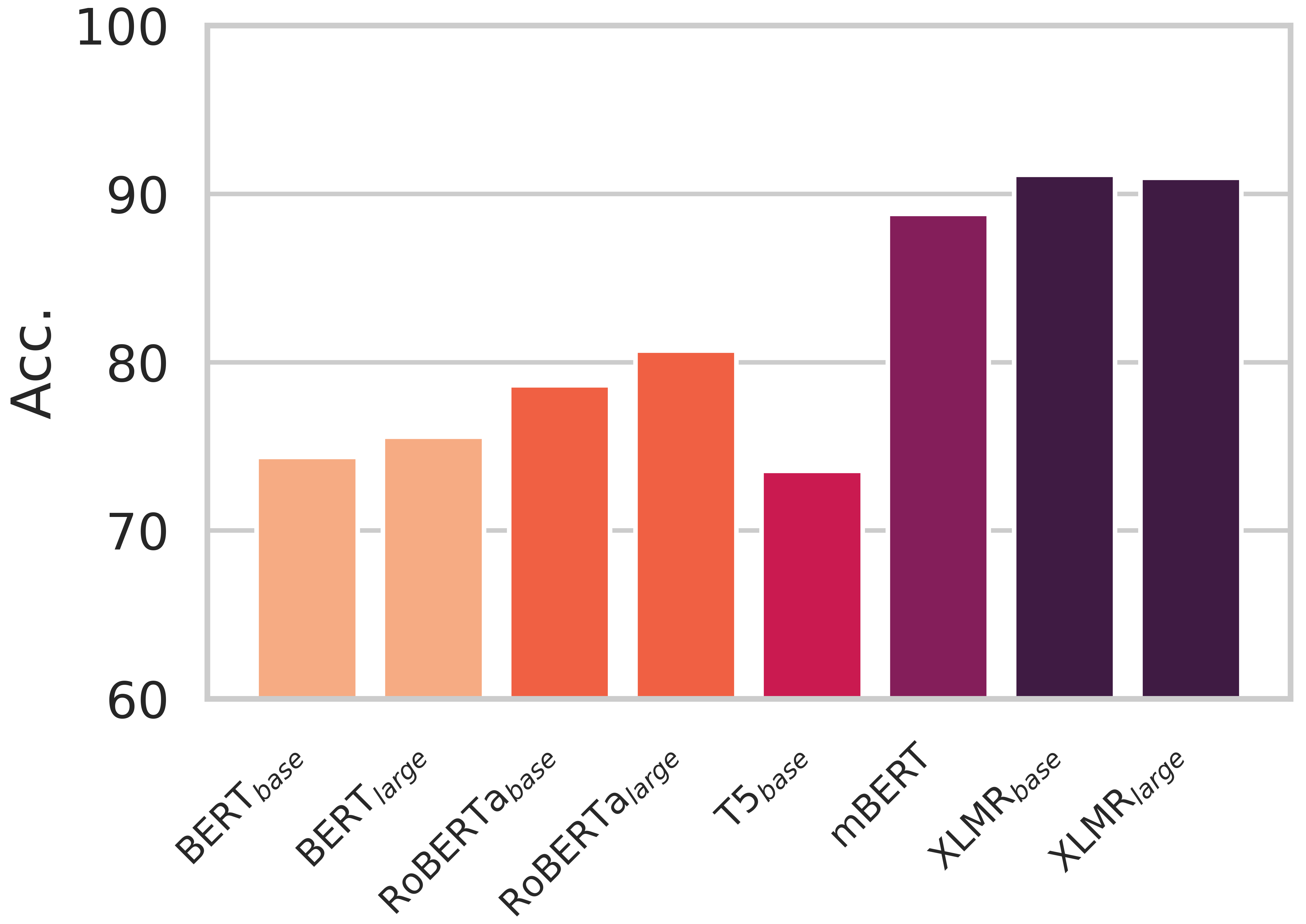} 
        \caption{POS (probing)}
        \label{fig:summary-2}
    \end{subfigure}
    \hfill
    \begin{subfigure}{0.32\textwidth}
        \centering
        \includegraphics[width=\linewidth]{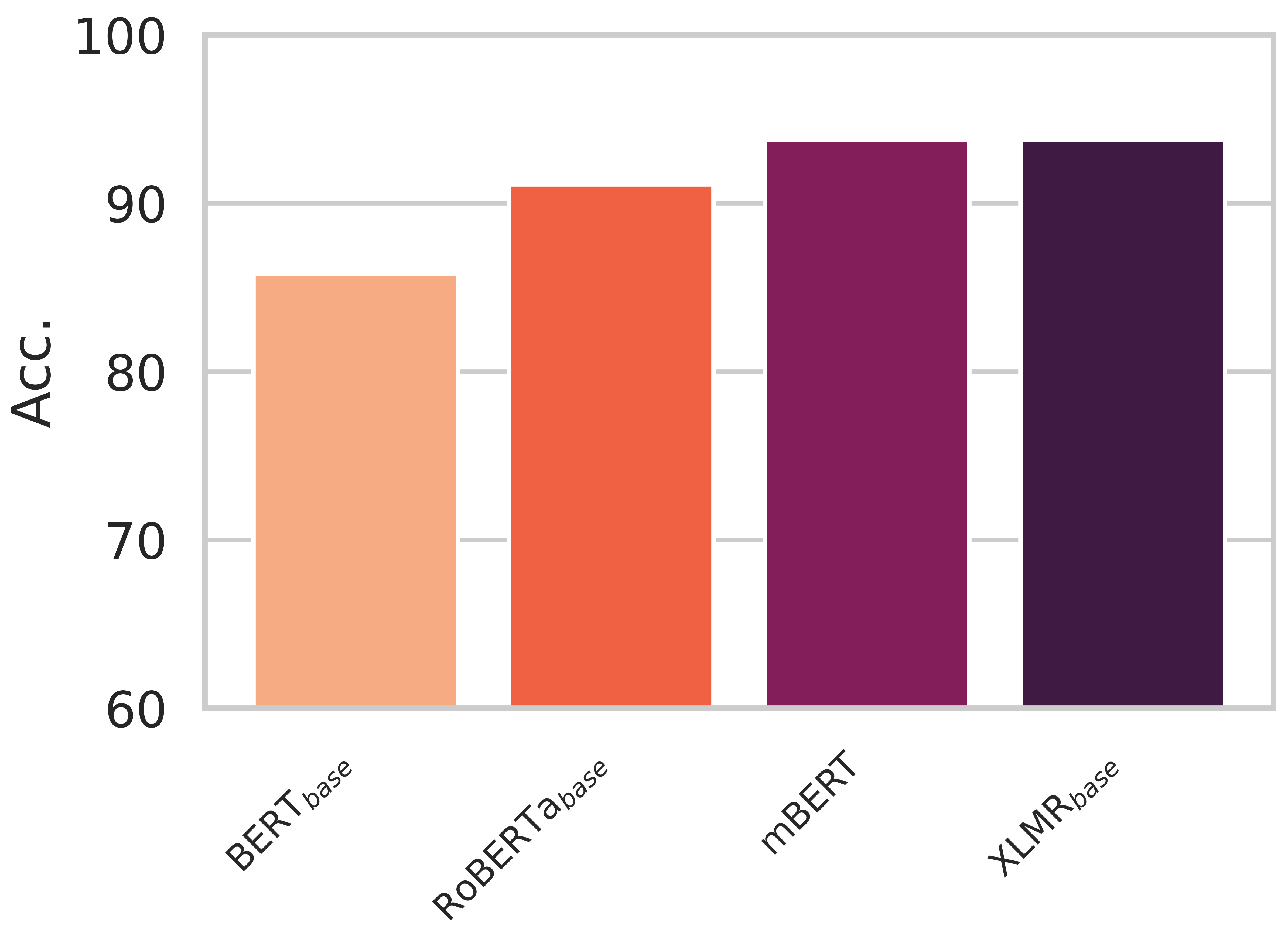} 
        \caption{POS (finetuned)}
        \label{fig:summary-3}
    \end{subfigure}
    \vspace{-5pt}
    \caption{Average performance by each model across all languages for the task. Lower is better for BPC.}
    \label{fig:exp-summary}
\end{figure*}

\subsection{Automatic Evaluation of Language Composition}
\label{subsection:auto-eval}
We use the FastText language identification model~\cite{joulin2017fasttext} to label every line in each corpus and keep lines as non-English if they score above a set confidence threshold (0.6). Due to the large size of \textsc{C4}, we subsample the first 50M examples (or 14\%); we classify the entirety of all other datasets. Since language detection is imperfect, particularly for low-resource languages \cite{caswell2021quality}, we present the results of this analysis as an estimate of the non-English data in each dataset and perform a qualitative analysis of potential errors in the following section.

A summary of the language identification experiments is presented in Figure \ref{fig:lang-comp-teaser}.\footnote{Full results of this evaluation are detailed in Appendix \ref{app:lang-id}.} We see that every corpus contains notable quantities of non-English data, with our estimates ranging between 300k to 406M tokens. An obvious factor that affects the amount of non-English data in each corpus is the overall size of the dataset; however, even when controlling for size by looking at the percentage of non-English data, we still see that the smaller corpora (\textsc{Wikipedia}, \textsc{BookCorpus}, and \textsc{Stories}) have relatively less non-English data. 

Indeed, a major factor of language leakage is the method in which the data was collected: the datasets derived from web crawls contain higher percentages of non-English text (\textsc{OpenWebText} and\textsc{CCNews}). This is true even for \textsc{C4}, where the dataset was filtered with a classifier to exclude non-English text \cite{raffel2020exploring}. Since automatic methods for language identification are imperfect, the datasets with more manual filtering (such as \textsc{Wikipedia}, which has human editors curating its content) are less prone to non-English data than those relying on classifiers. 
Due to these challenges, it is likely impossible to fully remove non-English text from a web-crawled dataset at scale. 

We also see that non-English text makes up small percentages of the overall data, though this still leads to millions of tokens in large datasets. 
The largest individual languages after English only make up 0.01\%, 0.15\%, and 0.05\% of the BERT, RoBERTa, and T5 training data, respectively.
Multilingual pretraining work has shown that models generalize to new languages from varying amounts of data \cite{delvin2019mbert, lample2019cross, conneau2020unsupervised}; however, these approaches intentionally select data across languages, and most upsample low-resource languages during training.
Without these considerations, it is an open question how well the models trained on these relatively small amounts of non-English data generalize.

\begin{table}[]
    \centering
    \scriptsize
    \begin{tabular}{c|c c c c c c}
        \toprule
        \multirow{2}{*}{\textbf{Type}} & \multicolumn{6}{c}{\textbf{Num. of Lines in...}} \\
         & \textbf{Book} & \textbf{Wiki} & \textbf{Stories} & \textbf{OpenWeb} & \textbf{CCNews} & \textbf{C4}\\
        \toprule
        \rowcolor{grey}
        & 156 & 129 & 99 & 175 & 193 & 169 \\
        \rowcolor{grey}
         & \multicolumn{6}{l}{\textbf{Ex:} Moraliska argument utgår ifrån våra moraliska intuitioner}  \\
        \rowcolor{grey}
        & \multicolumn{6}{l}{\qquad att rätt och fel inte endast är förankrade i människors vilja.}\\
        \rowcolor{grey}
        \multirow{-4}{*}{\textbf{NE}}& \multicolumn{6}{l}{\qquad (\textsc{OpenWebText})}\\
        & 13 & 11 & 15 & 4 & 1 & 22  \\
        & \multicolumn{6}{l}{\textbf{Ex:} The German blazon reads: "Von Silber über Schwarz }\\
        \multirow{-2}{*}{\textbf{BiL}} & \multicolumn{6}{l}{geteilt..." (\textsc{Wiki})}\\
        \rowcolor{grey}
        & 2 & 7 & 4 & 2 & 0 & 4 \\
        \rowcolor{grey}
         & \multicolumn{6}{l}{\textbf{Ex:} \textgreek{Εκείνη δεν μπορούσε να πληρώσει}}\\
        \rowcolor{grey}\multirow{-3}{*}{\textbf{Trans.}} & \multicolumn{6}{l}{\qquad  [She couldn’t pay.] (\textsc{BookCorpus})}\\
        & 1 & 28 & 5 & 1 & 0 & 1 \\
        & \multicolumn{6}{l}{\textbf{Ex:} 2012 Playhouse Presents \begin{CJK*}{UTF8}{min}ウィル シリーズ1、\end{CJK*}}\\
        \multirow{-3}{*}{\textbf{Ent.}} & \multicolumn{6}{l}{\qquad \begin{CJK*}{UTF8}{min}エピソード1:\end{CJK*} "The Minor Character" (\textsc{C4})}\\
        \toprule
        \rowcolor{grey}
        & 26 & 22 & 55 & 12 & 6 & 3 \\
        \rowcolor{grey}
        \multirow{-2}{*}{\textbf{En}} & \multicolumn{6}{l}{\textbf{Ex:} "Dere’s buzzards circlin’ ova dem trees." (\textsc{BookCorpus})}\\
        & 2 & 3 & 22 & 6 & 0 & 1 \\
        \multirow{-2}{*}{\textbf{XX}} & \multicolumn{6}{l}{\textbf{Ex:} M D | X O X | O O O = A (\textsc{Wiki})}\\
        \toprule
    \end{tabular}
    \caption{Results of the qualitative analysis of the non-English lines in various pretraining corpora. Type abbreviations are defined in Section \ref{subsection:qual-analysis}.}
    \label{tab:qual-analysis}
\end{table}

\subsection{Qualitative Analysis of Non-English Texts}
\label{subsection:qual-analysis}
We also perform a closer analysis on a random subset (200 per corpus) of non-English lines predicted by the language classifier (Table \ref{tab:qual-analysis}). Each example is manually coded into one of six categories. The first set covers various kinds of foreign language data: \textbf{NE}, where the line contains only \textit{non-English} language text; \textbf{BiL}, or \textit{bilingual}, where the line contains both English and non-English text; \textbf{Trans.}, in which the English and non-English data that are \textit{translations} of each other; and \textbf{Ent.}, where the line is primarily English but contains non-English \textit{entities}. The last two codes pertain to errors made by the language classifier: \textbf{En.}, where the line only contains \textit{English} text, and \textbf{XX}, which refers to lines that contain \textit{no natural language}.

The majority of lines across datasets consist only of non-English text. The next most common type of non-English data is \textbf{BiL}; this contains many subtypes of data, such as codeswitching and foreign language dialogue within English text. These datasets also include parallel data at both the sentence- and word-level.\footnote{e.g., "\begin{CJK*}{UTF8}{min}大学 【だい・がく】\end{CJK*}– college", \textsc{OpenWebText}}
We note that all observed translations are between English and another language.
Finally, some of the examples classified as non-English are actually English texts with non-English phrases.

Our analysis also shows that the language classifier performs worse on the non-web crawled data. For example, it misclassified a quarter of the sampled lines from \textsc{Stories} as non-English when they in fact only contain English text; many of these lines stem from snippets of dialogue in the dataset. We generally observe that lines coded as \textbf{En} tend to be shorter than the correctly labeled lines and often contain non-standard English. The language classifier also struggles to handle noisy lines, for which it has no appropriate language label.

\section{Cross-lingual Transfer of English Pretrained Models}
\label{sec:cross-ling-exps}

We now ask: how well do models pretrained on these putatively English corpora perform on non-English tasks? While the English data is more multilingual than previously thought, there are many differences between monolingual and multilingual pretraining; non-English data are often tokenized into more subword units\footnote{For example, the Basque UD treebank requires on average 1.78, 2.59, and 2.66 tokens per word to be encoded by XLMR, RoBERTa, and BERT, respectively.}
and are much less frequently observed during monolingual training. 

We evaluate popular English pretrained models on tasks in more than 50 languages: (masked) language modeling, POS probing, and finetuned POS tagging.
We compare the performance of monolingual BERT \cite{devlin2019bert}, RoBERTa \cite{liu2019roberta}, and T5 \cite{raffel2020exploring} against multilingual mBERT \cite{delvin2019mbert} and XLM-R \cite{conneau2020unsupervised}. We report average performance across five runs with different random seeds for the POS evaluations. The full results and all languages can be found in Appendix \ref{app:transfer}.

\subsection{Non-English MLM Evaluation}
We first measure the perplexity of English pretrained MLMs in other languages. We use Wiki-40B, a multilingual language modeling dataset that covers 41 languages \cite{guo2020wiki}. Following the Wiki-40B paper, we report bits per character (BPC) to allow comparison between models with different tokenizations of the text. 

We find that both BERT models perform notably worse on modeling other languages; however, RoBERTa, reduces the gap with the multilingual models from 2.51 BPC to 0.87 BPC (Figure \ref{fig:summary-1}). This finding is consistent with \citet{tran2020english}, who also found RoBERTa transfers well cross-lingually.

\subsection{POS Performance Across Languages}
Next, we evaluate how well monolingual English models perform on non-English downstream tasks, using part-of-speech (POS) tagging as a case study. 

\paragraph{Probing} We first consider the performance of the encoders when probed for POS knowledge (Figure \ref{fig:summary-2}).\footnote{For T5, this means that we evaluate the output of the encoder and discard the decoder.} Unsurprisingly, on average all of the English models underperform the multilingual models. Similar to MLM, we find that RoBERTa performs better than BERT when probed for POS features on other languages; surprisingly, it also strongly outperforms T5, despite \textsc{C4} containing more absolute non-English data than the RoBERTa corpus. 

This difference is likely due to two factors. First, in terms of relative percentages, RoBERTa is exposed to more non-English text than T5 (0.78\% compared to only 0.22\%). Secondly, RoBERTa's subword vocabulary is robust to unexpected inputs and does not substitute an UNK token any input tokens; in contrast, T5 and BERT have high rates of UNK tokens for some non-Latin languages (Appendix \ref{app:tokenization}).\footnote{\textit{UNK tokens} refer to placeholder tokens used when the model receives an input not covered by its vocabulary.}
However, for many high-resource languages the English models perform competitively, with T5 outperforming mBERT on German and Portuguese, among others.

\begin{table}[]
    \small
    \centering
    \addtolength{\tabcolsep}{-1.1pt}
    \begin{tabular}{c l | l l}
        \toprule
        \multirow{2}{*}{Task} & \multirow{2}{*}{Model} & \multicolumn{2}{c}{Corr. ($\rho$) with...}\\
        & & lang. data $\uparrow$ & en sim. $\downarrow$\\
        \toprule
        \multirow{4}{*}{\makecell{MLM\\(BPC) $\downarrow$}} 
        & BERT$_{base}$ & -0.258 & 0.097 \\
        & BERT$_{lg}$ & -0.258 & 0.118 \\
        & RoBERTa$_{base}$ & -0.667$^{**}$ & 0.326$^*$ \\
        & RoBERTa$_{lg}$ & -0.685$^{**}$ & 0.345$^*$ \\
        \hline 
        \multirow{5}{*}{\makecell{Frozen POS\\(Acc.) $\uparrow$}} 
         & BERT$_{base}$ & 0.335$^*$ & -0.332$^*$ \\
         & BERT$_{lg}$ & 0.314$^*$ & -0.375$^*$ \\
         & RoBERTa$_{base}$ & 0.594$^{**}$ & -0.260 \\
         & RoBERTa$_{lg}$ & 0.674$^{**}$ & -0.304$^*$ \\
         & T5$_{base}$ & 0.131 & -0.271 \\
        \hline 
        \multirow{2}{*}{\makecell{Finetuned POS\\ (Acc.) $\uparrow$}}
        & BERT$_{base}$ & 0.373$^*$ & -0.340$^*$ \\
        & RoBERTa$_{base}$ & 0.507$^{**}$ & -0.292$^*$ \\
        \toprule
    \end{tabular}
    \caption{Spearman correlations between task performance and (a) in-language data amounts in pretraining corpora (\textit{lang. data}) and (b) language similarity with English (\textit{en sim.}). $^{*} p<0.05$ and $^{**} p<0.001$.}
    \label{tab:ds_corrs}
\end{table}

\paragraph{Fine-tuning} To test if the effects of foreign language data carry through after finetuning, we also finetune a subset of the models (BERT$_{base}$, RoBERTa$_{base}$, mBERT, XLMR$_{base}$) for non-English POS tagging (Figure \ref{fig:summary-3}). After finetuning, the gap between the mono- and multilingual models is much smaller: RoBERTa only averages 2.65 points worse than XLM-R, compared to 12.5 points when probing.

\subsection{Potential Reasons for Cross-lingual Generalization}
We then investigate the correlation between potential transfer causes and model performance (Table \ref{tab:ds_corrs}). Specifically, we consider the quantity of target language data found in the model's pretraining corpus and the language similarity to English as potential causes of cross-lingual transfer.

We find that across tasks, RoBERTa task performance is most strongly correlated with the amount of target language data seen during pretraining. BERT and T5 task performance are less correlated with observed pretrained data, likely due to tokenization artifacts (Appendix \ref{app:tokenization}). Indeed, when we control for languages not written with Latin script on T5, the correlation between performance and the amount of target pretraining data increases to $\rho =$ 0.313.

We also consider the effect of language similarity on task performance, which is often hypothesized to facilitate cross-lingual transfer. We use the syntactic distance of languages calculated by \citet{malaviya17emnlp}; more similar languages score lower. However, we generally find that this is less correlated with performance than the quantity of target text, particularly for RoBERTa.

\section{Discussion}
In this paper, we demonstrate that English pretrained models are exposed to a considerable amount of non-English data during pretraining, particularly in the case of more recent models that are trained on larger corpora derived from web crawls. We also find that this non-English text acts as a significant source of signal for cross-lingual transfer.

Other recent work has focused on documenting the composition of pretraining corpora \cite{dodge2021documenting, gururangan2022whose}. \citet{caswell2021quality} manually audit a variety of multilingual datasets, finding data quality issues that are worse for low-resource languages and, similarly to our work, that texts for many languages are misclassified. 
In contrast, our focus is on the presence of foreign language data in primarily English corpora. 

Prior work has also shown the ability of monolingual models to transfer to other languages across a wide range of tasks \cite{gogoulou2021cross, li2021cross, tran2020english, artetxe2020cross, chi2020can}, but these works do not consider the effect of foreign language data leakage as a source of signal. Notably, \citet{de2021ability} mention the presence of foreign language data in their corpora but assume the small amounts observed will not affect model performance. However, our findings demonstrate that the amount of foreign language data directly correlates with cross-lingual transfer.


An obvious follow-up to our findings would be to retrain the models with text that is verified to only contain English data; this would confirm the effect the leaked non-English data has on the models. We reiterate that the standard method for filtering these datasets, automatic language classifiers, is imperfect. This, and the infeasibility of manual filtering due to the scale of the data, means that controlling for the language the model is pretrained on is nearly impossible.

However, the presence of foreign language data in pretraining corpora is not inherently problematic. Models trained on these datasets perform exceedingly well on their target languages \textit{and} generalize to other languages much better than expected. Rather, it is important to remember that these models are not performing zero-shot transfer when used in other languages, given the scale and data with which they were pretrained.

\section{Limitations}
Our work has a number of limitations. First, we measure the quantities of non-English data using a language classifier. The amounts of foreign language data we report are estimates for each dataset, as the classifier likely misclassified some examples. We manually audit the types of mistakes made by the language classifier in Section \ref{sec:lang-composition}. Additionally, we evaluate downstream performance via POS tagging, and it is possible that the models would exhibit different behavior on other NLP tasks.

We also only consider the effect of foreign language contamination for English pretrained models. It is unclear to what extent this phenomenon affects monolingual models for other languages; however, since many of the resources evaluated in this work are also used to pretrain non-English monolingual models (e.g., Wikipedia), similar effects would likely be observed.

\section*{Acknowledgements}
We would like to thank Hila Gonen, Julian Michael, and Ari Holtzman for their helpful conversations about the work. We also thank the anonymous reviewers for their thoughtful comments.

\bibliography{anthology,custom}
\bibliographystyle{acl_natbib}

\appendix

\section{Details of Transfer Experiments}
For the language modeling experiments, we perform whole word masking on 15\% of the words in the Wiki40B test set to calculate BPC. This experiment was zero-shot and required no further training of the models.

For the POS probing experiments, we train a linear classifier to predict POS from the final layer of each considered encoder; each probe therefore consists of a limited number of parameters $m*l$ where $m$ is the output dimension of the encoder being probed (768 for base models and 1024 for large models) and $l$ is the size of the label set (17 for POS tagging). For words that are tokenized into multiple subword units, we take the average representation of all tokens as the input to the classifier. When finetuning the model, we take the same setup as probing but unfreeze the encoder weights to allow them to update during training. The POS models are trained and evaluated on Universal Dependencies (UD) treebanks for each language \cite{nivre2020universal}.

We use a batch size of 256 for the frozen experiments and batch sizes of 16 for the finetuned models; we used a learning rate of 0.001 for the probing task and 5e-6 for finetuning. Due to the large number of experiments, we did not tune these parameters. For both POS tagging experiments, we use an Adam optimizer \cite{kingma2015adam}, and train each probe for 50 passes over the data (with early stopping on the validation set and a patience of 5). The pretrained models for all experiments are downloaded from Huggingface \cite{Wolf2019HuggingFacesTS}.

Each of our models was trained on a single Nvidia V100 GPU: 16GB for the frozen models and 32GB for the finetuned ones. The frozen probes each took between <1 and 8 minutes to train, and the finetuned probes were trained for between 5 minutes and 7.5 hours (depending on the dataset size, which varies by language, and early stopping epoch).

\section{The Effect of Tokenization}
\label{app:tokenization}

A factor that varies across the considered models is how they tokenize the input text for different languages. Table \ref{tab:tokens} gives the number of subword tokens per (white-space separated) word in the validation split of Wiki40b \cite{guo2020wiki}, as well as the percentage of tokens that are unked out by the tokenizer. We see that in general, all of the models (including explicitly multilingual ones) require more subword tokens per word for languages other than English.\footnote{We note that the number of subword tokens per ``word'' in Japanese is much larger than in other languages, as words in Japanese are not whitespace-separated.} We can also see that T5 is more efficient at encoding French, German, and Romanian than the other monolingual models (without a high UNK rate), likely because the T5 tokenizer was explicitly trained on English data mixed with those languages \cite{raffel2020exploring}. 

We also examine how many tokens are unked out by each tokenizer across languages. We see that BERT and T5 in particular have a high UNK rate ($>$ 10\%) for many languages not written in Latin script. This is in part due to the different tokenization schemes used by the models: RoBERTa uses a byte-level BPE encoding \cite{radford2019language}, which produces no UNK tokens for Unicode text, whereas the tokenization methods used by BERT and T5 (SentencePiece, \citet{kudo2018sentencepiece}) will unk out tokens not seen while training the tokenizer. 
Additionally, there are  other potential decisions made during tokenization that could affect these UNK rates, including filtering on non-Latin tokens or learning the subword tokenizer on a subset of the training data.

High UNK rates in the tokenized text for a language affect performance on downstream tasks. With regards to evaluating BPC, high frequencies of UNK tokens in the data likely make the language modeling task artificially easy, leading to lower BPC scores. Because of this, we note the cases where a model UNKs out more than 10\% of the considered data in the BPC results given in Table \ref{tab:full-bpc} with an asterisk (*). High UNK rates likely also lead to degraded performance on downstream tasks (including the considered POS tagging task in this work).

\section{Full Results of the Automatic Language Identity Analysis}
\label{app:lang-id}
We present a more complete set of results for the automatic language composition analysis (Section \ref{sec:lang-composition}) in Table \ref{tab:full-lang-composition}. We include every language that has 10,000 or more tokens in at least one of the considered corpora; we additionally report numbers for Basque and Frisian, as both languages are included in the experiments in Section \ref{sec:cross-ling-exps}.

\section{Full Results of Transfer Experiments}
\label{app:transfer}
Full results for whole word MLM are given in Table \ref{tab:full-bpc}; results for POS probing can be found in Table \ref{tab:full-frozen-pos} and results for finetuned POS tagging are detailed in Table \ref{tab:full-ft-pos}.

\input{tables/full_lang_comp_table}

\input{tables/tokenization_table}

\input{tables/full_bpc_table}

\input{tables/full_frozen_pos_table}

\input{tables/full_ft_pos_table}

\end{document}

%% file: tables/full_lang_comp_table.tex
\begin{table*}[]
    \centering
    \small
    \begin{tabular}{c l | r r r r r r | r r}
        \toprule
        \multirow{2}{*}{\textbf{ISO}} & \multirow{2}{*}{\textbf{Language}} & \multicolumn{8}{c}{\textbf{Number of Tokens}} \\
         & & \textbf{Wiki} & \textbf{Book} & \textbf{Stories} & \textbf{OpenWebText} & \textbf{CCNews} & \textbf{C4} & \textbf{BERT} & \textbf{RoBERTa} \\
         \hline
            \rowcolor{berry2}
            \textbf{en} & English & 2.0B & 802.4M & 6.2B & 6.4B & 13.0B & 17.8B & 2.8B & 28.3B \\
            \textbf{sq} & Albanian & 3.3K & 0 & 195 & 8.8K & 42.5M & 14.4K & 3.3K & 42.5M \\
            \rowcolor{grey}
            \textbf{es} & Spanish & 112.8K & 120.4K & 150.6K & 3.4M & 36.6M & 5.9M & 233.2K & 40.3M \\
            \textbf{de} & German & 176.2K & 5.4K & 104.8K & 3.1M & 34.4M & 9.0M & 181.5K & 37.8M \\
            \rowcolor{grey}
            \textbf{ro} & Romanian & 19.6K & 174 & 6.4K & 1.4M & 28.7M & 164.0K & 19.8K & 30.2M \\
            \textbf{pt} & Portugese & 43.2K & 760 & 44.2K & 1.5M & 10.0M & 1.9M & 44.0K & 11.5M \\
            \rowcolor{grey}
            \textbf{it} & Italian & 102.9K & 3.9K & 46.1K & 1.6M & 9.2M & 2.6M & 106.7K & 10.9M \\
            \textbf{fr} & French & 201.1K & 88.1K & 126.6K & 2.5M & 7.2M & 6.0M & 289.2K & 10.1M \\
            \rowcolor{grey}
            \textbf{pl} & Polish & 56.2K & 51 & 5.0K & 239.9K & 5.3M & 686.9K & 56.2K & 5.6M \\
            \textbf{nl} & Dutch & 28.0K & 1.0K & 37.3K & 254.7K & 4.4M & 1.7M & 29.0K & 4.8M \\
            \rowcolor{grey}
            \textbf{vi} & Vietnamese & 25.5K & 98 & 2.8K & 10.5K & 3.5M & 277.6K & 25.6K & 3.6M \\
            \textbf{tl} & Tagalog & 3.2K & 3.7K & 28.7K & 124.3K & 3.1M & 312.1K & 6.9K & 3.3M \\
            \rowcolor{grey}
            \textbf{cs} & Czech & 8.8K & 12 & 2.1K & 152.7K & 2.0M & 295.0K & 8.8K & 2.1M \\
            \textbf{fi} & Finnish & 6.9K & 119 & 4.7K & 243.2K & 1.7M & 214.5K & 7.0K & 1.9M \\
            \rowcolor{grey}
            \textbf{no} & Norwegian & 9.5K & 170 & 6.4K & 204.3K & 1.6M & 300.5K & 9.7K & 1.8M \\
            \textbf{hu} & Hungarian & 8.9K & 51 & 5.6K & 32.5K & 1.6M & 194.2K & 9.0K & 1.7M \\
            \rowcolor{grey}
            \textbf{hi} & Hindi & 6.7K & 0 & 520 & 32.2K & 1.5M & 328.0K & 6.7K & 1.6M \\
            \textbf{hr} & Croatian & 4.0K & 0 & 482 & 313.2K & 1.2M & 30.8K & 4.0K & 1.5M \\
            \rowcolor{grey}
            \textbf{id} & Indonesian & 1.5K & 100 & 12.9K & 83.5K & 1.3M & 997.7K & 1.6K & 1.4M \\
            \textbf{ru} & Russian & 17.4K & 606 & 3.9K & 956.3K & 64.8K & 2.3M & 18.0K & 1.0M \\
            \rowcolor{grey}
            \textbf{sv} & Swedish & 11.1K & 567 & 9.3K & 784.9K & 236.3K & 743.5K & 11.6K & 1.0M \\
            \textbf{sr} & Serbian & 753 & 0 & 709 & 39.0K & 976.2K & 36.8K & 753 & 1.0M \\
            \rowcolor{grey}
            \textbf{et} & Estonian & 2.8K & 0 & 288 & 8.0K & 817.1K & 32.0K & 2.8K & 828.2K \\
            \textbf{tr} & Turkish & 6.3K & 541 & 9.4K & 131.4K & 535.0K & 401.9K & 6.9K & 682.6K \\
            \rowcolor{grey}
            \textbf{af} & Afrikaans & 852 & 0 & 2.7K & 6.7K & 584.1K & 145.3K & 852 & 594.3K \\
            \textbf{ku} & Kurdish & 185 & 0 & 0 & 6.7K & 468.0K & 3.5K & 185 & 474.9K \\
            \rowcolor{grey}
            \textbf{da} & Danish & 3.1K & 20 & 5.3K & 249.8K & 157.8K & 271.1K & 3.1K & 415.9K \\
            \textbf{gl} & Galican & 101 & 0 & 309 & 637 & 317.5K & 9.6K & 101 & 318.6K \\
            \rowcolor{grey}
            \textbf{ja} & Japanese & 5.8K & 3.4K & 23.8K & 188.7K & 76.3K & 3.0M & 9.2K & 298.1K \\
            \textbf{ca} & Catalan & 5.2K & 99 & 418 & 28.2K & 258.8K & 108.3K & 5.3K & 292.8K \\
            \rowcolor{grey}
            \textbf{ar} & Arabic & 5.3K & 0 & 665 & 154.2K & 89.6K & 601.7K & 5.3K & 249.7K \\
            \textbf{ko} & Korean & 3.2K & 20 & 45 & 208.1K & 8.0K & 4.1M & 3.3K & 219.4K \\
            \rowcolor{grey}
            \textbf{el} & Greek & 15.2K & 777 & 1.8K & 123.8K & 28.4K & 288.7K & 16.0K & 169.9K \\
            \textbf{sl} & Slovenian & 262 & 0 & 250 & 102.1K & 14.5K & 46.8K & 262 & 117.1K \\
            \rowcolor{grey}
            \textbf{is} & Icelandic & 1.5K & 65.8K & 758 & 10.4K & 11.1K & 114.7K & 67.2K & 89.5K \\
            \textbf{ga} & Irish & 1.2K & 0 & 839 & 8.7K & 77.9K & 468.4K & 1.2K & 88.6K \\
            \rowcolor{grey}
            \textbf{uk} & Ukranian & 3.5K & 10 & 232 & 63.4K & 3.5K & 232.1K & 3.5K & 70.7K \\
            \textbf{he} & Hebrew & 5.2K & 0 & 4.6K & 46.5K & 9.6K & 138.0K & 5.2K & 66.0K \\
            \rowcolor{grey}
            \textbf{lt} & Lithuanian & 3.4K & 12 & 1.1K & 2.8K & 54.9K & 56.8K & 3.4K & 62.2K \\
            \textbf{sk} & Slovak & 1.9K & 0 & 76 & 16.2K & 43.9K & 64.7K & 1.9K & 62.0K \\
            \rowcolor{grey}
            \textbf{ms} & Malay & 896 & 29 & 1.4K & 1.8K & 45.9K & 42.8K & 925 & 50.0K \\
            \textbf{sw} & Swahili & 44 & 16 & 533 & 143 & 47.7K & 5.9K & 60 & 48.5K \\
            \rowcolor{grey}
            \textbf{eo} & Esperanto & 461 & 114 & 2.6K & 34.9K & 7.2K & 37.2K & 575 & 45.2K \\
            \textbf{zh} & Chinese & 4.1K & 12 & 5.5K & 30.8K & 4.5K & 410.2K & 4.1K & 44.8K \\
            \rowcolor{grey}
            \textbf{lv} & Latvian & 1.4K & 0 & 367 & 4.0K & 38.5K & 47.0K & 1.4K & 44.3K \\
            \textbf{bn} & Bengali & 2.5K & 24.8K & 51 & 6.2K & 10.1K & 48.6K & 27.3K & 43.6K \\
            \rowcolor{grey}
            \textbf{fa} & Persian & 3.0K & 0 & 261 & 28.2K & 5.8K & 668.9K & 3.0K & 37.2K \\
            \textbf{nn} & Norysk & 283 & 0 & 0 & 4.5K & 32.5K & 4.6K & 283 & 37.2K \\
            \rowcolor{grey}
            \textbf{la} & Latin & 6.0K & 641 & 3.8K & 19.4K & 4.7K & 34.7K & 6.6K & 34.5K \\
            \textbf{az} & Azerbaijani & 2.0K & 0 & 55 & 884 & 27.1K & 12.9K & 2.0K & 30.0K \\
            \rowcolor{grey}
            \textbf{th} & Thai & 7.6K & 0 & 592 & 15.1K & 5.5K & 131.8K & 7.6K & 28.7K \\
            \textbf{bg} & Bulgarian & 6.7K & 20 & 284 & 18.7K & 2.8K & 96.7K & 6.7K & 28.5K \\
            \rowcolor{grey}
            \textbf{cy} & Welsh & 1.2K & 84 & 440 & 18.5K & 7.4K & 59.5K & 1.3K & 27.6K \\
            \textbf{ilo} & Iloko & 19 & 0 & 16 & 628 & 18.8K & 1.1K & 19 & 19.5K \\
            \rowcolor{grey}
            \textbf{ur} & Urdu & 5.0K & 0 & 24 & 6.4K & 7.0K & 27.9K & 5.0K & 18.4K \\
            \textbf{ta} & Tamil & 5.4K & 0 & 234 & 5.9K & 5.8K & 42.2K & 5.4K & 17.4K \\
            \rowcolor{grey}
            \textbf{mt} & Maltese & 177 & 0 & 0 & 371 & 13.9K & 20.3K & 177 & 14.5K \\
            \textbf{hy} & Armenian & 2.6K & 0 & 0 & 7.5K & 2.9K & 21.5K & 2.6K & 13.0K \\
            \rowcolor{grey}
            \textbf{gd} & Gaelic & 198 & 0 & 52 & 874 & 8.8K & 117.6K & 198 & 10.0K \\
            \textbf{eu} & Basque & 99 & 5 & 1.8K & 2.8K & 2.4K & 19.3K & 104 & 7.0K \\
            \rowcolor{grey}
            \textbf{fy} & Frisian & 80 & 0 & 1.3K & 1.5K & 601 & 9.8K & 80 & 3.4K \\
        \hline
        \multicolumn{2}{l|}{\textbf{Total (Non-En)}} & 983k & 322k & 682k & 18.6M & 201M & 406M$^\dagger$ & 1.3M & 222M \\
        \toprule
    \end{tabular}
    \caption{Full results for the automatic language composition analysis of pretraining corpora presented in Section \ref{sec:lang-composition}. The last two columns include the total data that BERT and RoBERTa were trained on, respectively; C4 contains the data T5 was trained on, and contains the estimates for the first 50M examples in the full C4 dataset; $\dagger$ represents the projected estimate for the full dataset.}
    \label{tab:full-lang-composition}
\end{table*}

%% file: tables/tokenization_table.tex
\begin{table*}[]
    \centering
    \begin{tabular}{c | r r r | r r}
        \toprule
        \multirow{2}{*}{\textbf{ISO}} & \multicolumn{3}{c}{\textbf{Monolingual}} & \multicolumn{2}{c}{\textbf{Multilingual}}  \\
         & \makecell[c]{\textbf{BERT}} & \makecell[c]{\textbf{RoBERTa}} & \makecell[c]{\textbf{T5}} & \makecell[c]{\textbf{mBERT}} & \makecell[c]{\textbf{XLMR}} \\
        \toprule
        \textbf{ar} & 2.91 (0.60\%) & 3.11 (0.0\%) & 1.91 (\textbf{41.26\%}) & 1.95 (0.05\%) & 1.73 (9.8e-4\%) \\
        \rowcolor{grey}
        \textbf{bg} & 3.03 (0.06\%) & 3.25 (0.0\%) & 2.88 (\textbf{17.83\%}) & 1.93 (0.67\%) & 1.72 (1.2e-3\%) \\
        \textbf{ca} & 1.95 (0.01\%) & 1.83 (0.0\%) & 2.08 (1.23\%) & 1.55 (0.12\%) & 1.56 (4.6e-4\%) \\
        \rowcolor{grey}
        \textbf{cs} & 2.64 (0.03\%) & 2.64 (0.0\%) & 2.85 (\textbf{10.30\%}) & 2.00 (0.22\%) & 1.86 (5.6e-4\%) \\
        \textbf{da} & 2.19 (0.01\%) & 2.07 (0.0\%) & 2.42 (3.56\%) & 1.74 (0.14\%) & 1.63 (4.5e-4\%) \\
        \rowcolor{grey}
        \textbf{de} & 2.21 (0.02\%) & 2.14 (0.0\%) & 1.85 (0.26\%) & 1.65 (0.37\%) & 1.67 (1.5e-3\%)  \\
        \textbf{el} & 2.74 (0.36\%) & 2.96 (0.0\%) & 1.82 (\textbf{40.30\%}) & 2.05 (0.05\%) & 1.73 (1.4e-3\%)  \\
        \rowcolor{berry2}
        \textbf{en} & 1.38 (0.03\%) & 1.32 (0.0\%) & 1.44 (0.15\%) & 1.37 (0.22\%) & 1.42 (1.3e-3\%) \\
        \textbf{es} & 1.76 (0.01\%) & 1.67 (0.0\%) & 1.88 (1.48\%) & 1.37 (0.11\%) & 1.37 (8.2e-4\%) \\
        \rowcolor{grey}
        \textbf{et} & 3.02 (0.03\%) & 2.92 (0.0\%) & 3.35 (1.55\%) & 2.47 (0.33\%) & 2.21 (1.5e-3\%) \\
        \textbf{fa} & 2.80 (1.12\%) & 3.34 (0.0\%) & 2.00 (\textbf{42.14\%}) & 1.70 (0.05\%) & 1.55 (6.1e-3\%) \\
        \rowcolor{grey}
        \textbf{fi} & 3.18 (8.3e-3\%) & 3.06 (0.0\%) & 3.48 (0.13\%) & 2.45 (0.35\%) & 2.24 (9.9e-4\%) \\
        \textbf{fr} & 1.90 (0.01\%) & 1.81 (0.0\%) & 1.78 (0.26\%) & 1.53 (0.43\%) & 1.57 (8.5e-4\%)  \\
        \rowcolor{grey}
        \textbf{he} & 2.82 (0.72\%) & 3.08 (0.0\%) & 1.97 (\textbf{40.30\%}) & 2.05 (0.06\%) & 1.89 (4.2e-4\%) \\
        \textbf{hi} & 1.98 (\textbf{12.06\%}) & 2.84 (0.0\%) & 1.64 (\textbf{42.82\%}) & 1.64 (0.05\%) & 1.39 (1.3e-3\%) \\
        \rowcolor{grey}
        \textbf{hr} & 2.38 (7.1e-3\%) & 2.27 (0.0\%) & 2.57 (3.71\%) & 1.85 (0.08\%) & 1.73 (1.1e-3\%) \\
        \textbf{hu} & 2.78 (0.02\%) & 2.72 (0.0\%) & 3.00 (2.60\%) & 2.12 (0.28\%) & 1.93 (1.1e-3\%) \\
        \rowcolor{grey}
        \textbf{id} & 2.34 (0.05\%) & 2.22 (0.0\%) & 2.54 (0.13\%) & 1.70 (0.12\%) & 1.59 (4.5e-3\%) \\
        \textbf{it} & 1.92 (0.01\%) & 1.83 (0.0\%) & 2.05 (0.42\%) & 1.51 (0.10\%) & 1.52 (1.0e-3\%) \\
        \rowcolor{grey}
        \textbf{ja} & 34.41 (\textbf{39.97\%}) & 47.67 (0.0\%) & 9.50 (\textbf{22.19\%}) & 35.22 (0.05\%) & 31.30 (0.03\%) \\
        \textbf{ko} & 1.60 (\textbf{59.65\%}) & 4.78 (0.0\%) & 2.32 (\textbf{38.63\%}) & 2.65 (0.25\%) & 2.53 (0.03\%) \\
        \rowcolor{grey}
        \textbf{lt} & 3.06 (0.80\%) & 3.12 (0.0\%) & 3.41 (8.78\%) & 2.48 (0.97\%) & 2.23 (7.7e-3\%) \\
        \textbf{lv} & 2.89 (0.48\%) & 2.84 (0.0\%) & 3.13 (\textbf{12.52\%}) & 2.36 (0.30\%) & 2.06 (2.8e-3\%) \\
        \rowcolor{grey}
        \textbf{ms} & 2.34 (0.03\%) & 2.21 (0.0\%) & 2.53 (0.10\%) & 1.71 (0.10\%) & 1.58 (1.9e-3\%) \\
        \textbf{nl} & 2.18 (8.0e-3\%) & 2.04 (0.0\%) & 2.31 (0.21\%) & 1.64 (0.08\%) & 1.62 (4.6e-4\%)\\
        \rowcolor{grey}
        \textbf{no} & 2.24 (0.03\%) & 2.10 (0.0\%) & 2.49 (3.01\%) & 1.74 (0.13\%) & 1.66 (2.0e-3\%) \\
        \textbf{pl} & 2.60 (9.3e-3\%) & 2.60 (0.0\%) & 2.83 (6.50\%) & 1.96 (0.44\%) & 1.88 (7.2e-4\%) \\
        \rowcolor{grey}
        \textbf{pt} & 1.86 (0.02\%) & 1.76 (0.0\%) & 2.01 (2.05\%) & 1.45 (0.11\%) & 1.43 (1.0-3\%) \\
        \textbf{ro} & 2.03 (0.01\%) & 2.02 (0.0\%) & 1.73 (0.18\%) & 1.63 (0.25\%) & 1.54 (7.9e-4\%) \\
        \rowcolor{grey}
        \textbf{ru} & 3.05 (0.02\%) & 3.25 (0.0\%) & 2.90 (\textbf{21.1\%}) & 1.92 (0.53\%) & 1.82 (2.1e-3\%) \\
        \textbf{sk} & 2.86 (0.05\%) & 2.81 (0.0\%) & 3.14 (7.08\%) & 2.20 (0.19\%) & 2.00 (1.4e-3\%) \\
        \rowcolor{grey}
        \textbf{sl} & 2.37 (9.7e-3\%) & 2.24 (0.0\%) & 2.53 (3.45\%) & 1.91 (0.06\%) & 1.73 (1.1e-3\%) \\
        \textbf{sr} & 3.01 (0.71\%) & 3.33 (0.0\%) & 2.95 (\textbf{17.14\%}) & 1.95 (0.19\%) & 1.77 (4.8e-4\%) \\
        \rowcolor{grey}
        \textbf{sv} & 2.57 (7.9e-3\%) & 2.40 (0.0\%) & 2.77 (2.08\%) & 1.90 (0.15\%) & 1.80 (8.5e-4\%) \\
        \textbf{th} & 2.13 (\textbf{36.91\%}) & 11.79 (0.0\%) & 2.73 (\textbf{28.58\%}) & 8.34 (0.12\%) & 5.42 (1.6e-3\%) \\
        \rowcolor{grey}
        \textbf{tl} & 2.14 (0.10\%) & 2.02 (0.0\%) & 2.44 (0.18\%) & 1.81 (0.12\%) & 1.70 (2.9e-3\%) \\
        \textbf{tr} & 2.94 (0.01\%) & 2.87 (0.0\%) & 3.19 (7.36\%) & 2.13 (0.31\%) & 1.91 (2.0e-3\%) \\
        \rowcolor{grey}
        \textbf{uk} & 3.36 (0.52\%) & 3.73 (0.0\%) & 3.23 (\textbf{24.12\%}) & 2.11 (0.54\%) & 1.94 (1.4e-3\%) \\
        \textbf{vi} & 1.76 (1.44\%) & 1.95 (0.0\%) & 1.89 (\textbf{15.12\%}) & 1.19 (0.08\%) & 1.16 (3.2e-3\%) \\ 
        \toprule
    \end{tabular}
    \caption{The average number of subword tokens per white-spaced word (and the percentage of UNKed out tokens) in the Wiki40b validation set for each language. Cases where more than 10\% of tokens are unked out are in bold.} 
    \label{tab:tokens}
\end{table*}

%% file: tables/full_bpc_table.tex
\begin{table*}[]
    \centering
    \begin{tabular}{c | c c c c | c c c}
        \toprule
        \multirow{2}{*}{\textbf{ISO}} & \multicolumn{4}{c}{\textbf{Monolingual}} & \multicolumn{3}{c}{\textbf{Multilingual}}  \\
         & \textbf{BERT$_{ba}$} & \textbf{BERT$_{lg}$} & \textbf{RoBERTa$_{ba}$} & \textbf{RoBERTa$_{lg}$} & \textbf{mBERT} & \textbf{XLMR$_{ba}$} & \textbf{XLMR$_{lg}$} \\
        \toprule
        \textbf{ar} & 6.214 & 9.331 & 3.319 & 3.899 & 1.849 & 1.871 & 1.691 \\
        \rowcolor{grey}
        \textbf{bg} & 6.334 & 7.883 & 3.544 & 3.587 & 1.553 & 1.494 & 1.358 \\
        \textbf{ca} & 3.382 & 3.565 & 1.834 & 1.640 & 1.108 & 1.477 & 1.329 \\
        \rowcolor{grey}
        \textbf{cs} & 4.316 & 4.738 & 2.634 & 2.493 & 1.703 & 1.715 & 1.533 \\
        \textbf{da} & 3.560 & 3.832 & 2.104 & 1.931 & 1.420 & 1.427 & 1.272 \\
        \rowcolor{grey}
        \textbf{de} & 3.430 & 3.644 & 1.815 & 1.634 & 1.102 & 1.361 & 1.218 \\
        \textbf{el} & 6.934 & 8.915 & 3.852 & 3.885 & 1.793 & 1.588 & 1.440 \\
        \rowcolor{berry2}
        \textbf{en} & 1.285 & 1.377 & 0.595 & 0.516 & 0.938 & 1.249 & 1.131 \\
        \textbf{es} & 3.281 & 3.551 & 1.526 & 1.345 & 1.036 & 1.284 & 1.165 \\
        \rowcolor{grey}
        \textbf{et} & 3.846 & 4.108 & 2.448 & 2.318 & 1.878 & 1.858 & 1.671 \\
        \textbf{fa} & 5.813 & 8.501 & 3.614 & 4.113 & 1.723 & 1.567 & 1.418 \\
        \rowcolor{grey}
        \textbf{fi} & 3.732 & 4.064 & 2.357 & 2.240 & 1.633 & 1.618 & 1.451 \\
        \textbf{fr} & 3.213 & 3.439 & 1.586 & 1.414 & 1.038 & 1.434 & 1.305 \\
        \rowcolor{grey}
        \textbf{he} & 6.490 & 9.074 & 3.530 & 3.831 & 1.817 & 1.976 & 1.739 \\
        \textbf{hi} & 4.240* & 5.503* & 1.487 & 1.407 & 1.876 & 1.641 & 1.516 \\
        \rowcolor{grey}
        \textbf{hr} & 3.972 & 4.298 & 2.267 & 2.109 & 1.563 & 1.644 & 1.484 \\
        \textbf{hu} & 4.203 & 4.585 & 2.741 & 2.632 & 1.778 & 1.713 & 1.548 \\
        \rowcolor{grey}
        \textbf{id} & 3.436 & 3.665 & 1.976 & 1.838 & 1.221 & 1.243 & 1.129 \\
        \textbf{it} & 3.263 & 3.536 & 1.661 & 1.475 & 1.098 & 1.402 & 1.256 \\
        \rowcolor{grey}
        \textbf{ja} & 1.840* & 2.065* & 5.481 & 6.775 & 2.082 & 6.827 & 8.016 \\
        \textbf{ko} & 0.781* & 0.846* & 4.204 & 4.639 & 3.144 & 3.504 & 3.241 \\
        \rowcolor{grey}
        \textbf{lt} & 3.953 & 4.271 & 2.746 & 2.633 & 1.840 & 1.789 & 1.604 \\
        \textbf{lv} & 4.231 & 4.512 & 2.833 & 2.730 & 1.890 & 1.750 & 1.548 \\
        \rowcolor{grey}
        \textbf{ms} & 3.461 & 3.698 & 2.010 & 1.886 & 1.280 & 1.365 & 1.257 \\
        \textbf{nl} & 3.445 & 3.693 & 1.855 & 1.680 & 1.222 & 1.397 & 1.257 \\
        \rowcolor{grey}
        \textbf{no} & 3.580 & 3.873 & 2.052 & 1.872 & 1.398 & 1.469 & 1.312 \\
        \textbf{pl} & 4.020 & 4.505 & 2.506 & 2.365 & 1.495 & 1.604 & 1.437 \\
        \rowcolor{grey}
        \textbf{pt} & 3.442 & 3.718 & 1.658 & 1.465 & 1.128 & 1.316 & 1.190 \\
        \textbf{ro} & 3.641 & 3.929 & 1.950 & 1.772 & 1.402 & 1.435 & 1.286 \\
        \rowcolor{grey}
        \textbf{ru} & 6.747 & 8.122 & 3.624 & 3.673 & 1.385 & 1.491 & 1.344 \\
        \textbf{sk} & 4.263 & 4.628 & 2.714 & 2.594 & 1.804 & 1.753 & 1.594 \\
        \rowcolor{grey}
        \textbf{sl} & 3.972 & 4.294 & 2.415 & 2.273 & 1.642 & 1.563 & 1.391 \\
        \textbf{sr} & 6.081 & 7.216 & 3.610 & 3.661 & 1.772 & 1.783 & 1.681 \\
        \rowcolor{grey}
        \textbf{sv} & 3.774 & 4.081 & 2.196 & 2.019 & 1.460 & 1.523 & 1.372 \\
        \textbf{th} & 1.551* & 1.689* & 3.312 & 3.535 & 3.861 & 2.119 & 2.237 \\
        \rowcolor{grey}
        \textbf{tl} & 3.250 & 3.458 & 1.763 & 1.623 & 1.616 & 1.713 & 1.572 \\
        \textbf{tr} & 4.102 & 4.427 & 2.715 & 2.585 & 1.635 & 1.603 & 1.460 \\
        \rowcolor{grey}
        \textbf{uk} & 6.542 & 7.912 & 3.763 & 3.823 & 1.566 & 1.635 & 1.488 \\
        \textbf{vi} & 5.134 & 5.794 & 2.590 & 2.574 & 1.046 & 1.191 & 1.055 \\ 
        \toprule
    \end{tabular}
    \caption{Full results for the zero-shot BPC experiments in Section \ref{sec:cross-ling-exps}. Results noted with * correspond to cases of high UNK rates in the tokenization of the data (Section \ref{app:tokenization}).}
    \label{tab:full-bpc}
\end{table*}

%% file: tables/full_frozen_pos_table.tex
\begin{table*}[]
    \centering
    \fontsize{8.5pt}{10.2pt}\selectfont
    \begin{tabular}{c | c c | c c c c c | c c c}
        \toprule
        \multirow{2}{*}{\textbf{ISO}} & \multicolumn{3}{c}{\textbf{Baselines}} & \multicolumn{4}{c}{\textbf{Monolingual}} & \multicolumn{3}{c}{\textbf{Multilingual}}  \\
         & \textbf{Maj. Label} & \textbf{Word Maj.} & \textbf{BERT$_{ba}$} & \textbf{BERT$_{lg}$} & \textbf{Ro$_{ba}$} & \textbf{Ro$_{lg}$} & \textbf{T5-base} & \textbf{mBERT} & \textbf{XLMR$_{ba}$} & \textbf{XLMR$_{lg}$} \\
        \toprule
\textbf{af} & 21.650 & 83.335 & 81.695 & 84.855 & 88.858 & 92.324 & 90.464 & 93.590 & 97.490 & 96.381 \\
\rowcolor{grey}
\textbf{ar} & 33.297 & 90.148 & 79.595 & 79.994 & 78.939 & 79.242 & 43.182 & 93.724 & 95.659 & 95.533 \\
\textbf{bg} & 21.834 & 86.091 & 85.617 & 84.238 & 78.514 & 81.147 & 85.304 & 94.977 & 97.240 & 97.103 \\
\rowcolor{grey}
\textbf{ca} & 17.868 & 90.984 & 93.208 & 93.432 & 94.333 & 94.545 & 95.863 & 97.610 & 98.222 & 98.202 \\
\textbf{cs} & 24.708 & 91.284 & 82.312 & 80.591 & 89.272 & 93.488 & 86.560 & 96.554 & 97.548 & 97.744 \\
\rowcolor{grey}
\textbf{cy} & 31.099 & 73.587 & 69.625 & 72.641 & 69.171 & 70.309 & 76.220 & 81.713 & 76.690 & 77.215 \\
\textbf{da} & 18.606 & 77.841 & 81.137 & 81.485 & 85.308 & 91.057 & 86.832 & 91.901 & 96.757 & 96.087 \\
\rowcolor{grey}
\textbf{de} & 17.784 & 81.992 & 86.663 & 88.306 & 91.266 & 93.074 & 92.878 & 92.022 & 94.851 & 94.020 \\
\textbf{el} & 21.148 & 81.128 & 79.996 & 79.110 & 66.604 & 76.795 & 37.050 & 92.509 & 95.435 & 95.371 \\
\rowcolor{berry2}
\textbf{en} & 16.999 & 82.920 & 93.803 & 92.903 & 94.785 & 94.418 & 96.286 & 93.666 & 95.366 & 95.342 \\
\textbf{es} & 17.734 & 90.734 & 89.639 & 92.860 & 97.652 & 93.171 & 97.222 & 97.699 & 98.418 & 98.497 \\
\rowcolor{grey}
\textbf{et} & 26.462 & 78.486 & 74.987 & 78.524 & 74.460 & 81.150 & 79.287 & 91.891 & 90.717 & 91.143 \\
\textbf{eu} & 24.422 & 77.591 & 73.152 & 75.663 & 72.713 & 72.254 & 80.294 & 84.712 & 88.744 & 88.023 \\
\rowcolor{grey}
\textbf{fa} & 33.521 & 91.916 & 78.545 & 78.202 & 67.668 & 66.767 & 46.485 & 93.025 & 96.310 & 96.597 \\
\textbf{fi} & 27.965 & 74.378 & 71.083 & 72.301 & 77.318 & 82.587 & 77.630 & 92.621 & 95.823 & 95.872 \\
\rowcolor{grey}
\textbf{fr} & 18.749 & 89.584 & 90.960 & 90.197 & 93.690 & 95.278 & 96.612 & 95.963 & 95.837 & 95.813 \\
\textbf{fy} & 14.815 & 85.190 & 79.749 & 82.128 & 78.766 & 78.216 & 86.351 & 90.898 & 87.908 & 89.405 \\
\rowcolor{grey}
\textbf{ga} & 29.122 & 81.512 & 72.470 & 77.009 & 76.983 & 79.169 & 82.430 & 87.097 & 91.493 & 92.583 \\
\textbf{gd} & 21.166 & 80.114 & 78.264 & 79.641 & 74.989 & 76.281 & 79.590 & 78.387 & 84.102 & 85.609 \\
\rowcolor{grey}
\textbf{gl} & 22.969 & 86.294 & 87.727 & 89.058 & 92.638 & 93.176 & 93.647 & 92.559 & 95.145 & 95.548 \\
\textbf{he} & 23.601 & 85.491 & 75.040 & 75.393 & 69.930 & 70.160 & 45.758 & 93.206 & 96.403 & 94.864 \\
\rowcolor{grey}
\textbf{hi} & 22.128 & 89.365 & 68.650 & 68.861 & 77.250 & 80.597 & 38.185 & 94.054 & 95.524 & 94.250 \\
\textbf{hr} & 24.182 & 83.533 & 80.408 & 82.370 & 92.955 & 94.742 & 87.784 & 96.164 & 98.011 & 98.313 \\
\rowcolor{grey}
\textbf{hu} & 22.429 & 60.356 & 72.619 & 73.444 & 76.403 & 83.633 & 79.868 & 88.273 & 93.404 & 91.868 \\
\textbf{hy} & 24.995 & 68.931 & 50.956 & 52.033 & 58.560 & 58.792 & 44.142 & 89.001 & 90.403 & 93.937 \\
\rowcolor{grey}
\textbf{id} & 21.642 & 81.278 & 4.151 & 4.151 & 79.539 & 81.664 & 4.151 & 4.151 & 4.151 & 4.151 \\
\textbf{is} & 17.286 & 90.407 & 80.969 & 84.217 & 79.792 & 82.154 & 84.121 & 91.414 & 97.459 & 97.778 \\
\rowcolor{grey}
\textbf{it} & 19.920 & 89.758 & 89.497 & 91.317 & 93.991 & 95.521 & 94.715 & 96.662 & 97.454 & 96.854 \\
\textbf{ja} & 30.137 & 85.592 & 76.268 & 75.659 & 83.491 & 85.013 & 39.409 & 92.362 & 90.844 & 91.080 \\
\rowcolor{grey}
\textbf{ko} & 30.011 & 67.715 & 48.090 & 47.683 & 67.852 & 70.745 & 47.288 & 76.359 & 80.372 & 80.704 \\
\textbf{la} & 21.355 & 94.888 & 91.416 & 94.079 & 92.133 & 92.489 & 95.928 & 95.008 & 98.250 & 97.548 \\
\rowcolor{grey}
\textbf{lt} & 31.345 & 61.009 & 67.026 & 70.382 & 67.892 & 66.602 & 77.164 & 90.542 & 91.641 & 94.710 \\
\textbf{lv} & 27.108 & 79.198 & 71.889 & 77.190 & 72.061 & 74.857 & 81.630 & 87.627 & 93.565 & 92.775 \\
\rowcolor{grey}
\textbf{mt} & 19.489 & 76.131 & 75.582 & 78.313 & 75.199 & 75.156 & 80.854 & 76.877 & 70.113 & 74.191 \\
\textbf{nl} & 16.799 & 81.878 & 79.448 & 84.540 & 90.126 & 92.816 & 89.207 & 94.356 & 95.920 & 95.990 \\
\rowcolor{grey}
\textbf{pl} & 24.900 & 83.868 & 82.357 & 81.721 & 91.563 & 92.491 & 90.494 & 94.184 & 98.231 & 97.840 \\
\textbf{pt} & 18.117 & 83.517 & 85.594 & 86.801 & 91.591 & 92.114 & 95.120 & 94.066 & 96.981 & 94.810 \\
\rowcolor{grey}
\textbf{ro} & 24.849 & 85.537 & 82.630 & 84.414 & 93.830 & 95.332 & 93.407 & 95.284 & 97.443 & 97.229 \\
\textbf{ru} & 23.843 & 88.593 & 84.258 & 85.526 & 82.713 & 86.811 & 88.812 & 95.299 & 96.943 & 94.714 \\
\rowcolor{grey}
\textbf{sk} & 19.264 & 61.821 & 80.441 & 83.025 & 86.850 & 89.339 & 86.872 & 92.414 & 96.290 & 95.810 \\
\textbf{sl} & 21.289 & 77.815 & 82.459 & 80.959 & 88.357 & 88.176 & 87.189 & 96.388 & 97.953 & 98.144 \\
\rowcolor{grey}
\textbf{sr} & 24.378 & 82.523 & 84.930 & 85.434 & 94.762 & 94.769 & 89.884 & 92.727 & 98.638 & 98.333 \\
\textbf{sv} & 17.579 & 78.993 & 71.818 & 79.061 & 78.523 & 88.662 & 83.226 & 92.826 & 96.246 & 95.350 \\
\rowcolor{grey}
\textbf{ta} & 29.389 & 53.042 & 43.992 & 40.563 & 38.361 & 43.228 & 43.147 & 74.912 & 76.521 & 75.606 \\
\textbf{tr} & 36.494 & 82.343 & 78.115 & 73.015 & 67.656 & 67.340 & 80.830 & 88.633 & 91.392 & 89.939 \\
\rowcolor{grey}
\textbf{uk} & 23.213 & 71.964 & 78.214 & 78.950 & 65.442 & 69.780 & 80.780 & 91.836 & 96.213 & 96.823 \\
\textbf{ur} & 23.564 & 85.736 & 68.112 & 68.819 & 69.202 & 67.458 & 34.116 & 88.954 & 91.170 & 92.341 \\
\rowcolor{grey}
\textbf{vi} & 32.019 & 75.901 & 54.764 & 54.267 & 54.054 & 56.402 & 60.768 & 75.844 & 85.855 & 81.554 \\
\textbf{zh} & 27.478 & 78.696 & 49.462 & 51.400 & 64.537 & 67.238 & 44.479 & 87.363 & 88.565 & 85.921 \\
        \toprule
    \end{tabular}
    \caption{Full results for the frozen POS tagging experiments in Section \ref{sec:cross-ling-exps}. }
    \label{tab:full-frozen-pos}
\end{table*}

%% file: tables/full_ft_pos_table.tex
\begin{table*}[]
    \centering
    \fontsize{8.5pt}{10.2pt}\selectfont
    \begin{tabular}{c | c  c |c c}
        \toprule
        \multirow{2}{*}{\textbf{ISO}} & \multicolumn{2}{c}{\textbf{Monolingual}} & \multicolumn{2}{c}{\textbf{Multilingual}}  \\
         & \textbf{BERT$_{ba}$} & \textbf{Ro$_{ba}$} &  \textbf{mBERT} & \textbf{XLMR$_{ba}$} \\
        \toprule
\textbf{af} & 92.108 & 94.528 & 96.802 & 97.158 \\
\rowcolor{grey}
\textbf{ar} & 92.896 & 93.417 & 96.151 & 96.673 \\
\textbf{bg} & 97.185 & 96.797 & 98.714 & 99.145 \\
\rowcolor{grey}
\textbf{ca} & 98.058 & 98.264 & 98.813 & 98.845 \\
\textbf{cs} & 98.191 & 98.306 & 98.803 & 98.952 \\
\rowcolor{grey}
\textbf{cy} & 82.152 & 72.484 & 92.109 & 84.927 \\
\textbf{da} & 93.134 & 93.872 & 97.037 & 97.556 \\
\rowcolor{grey}
\textbf{de} & 93.285 & 93.554 & 95.178 & 95.189 \\
\textbf{el} & 92.725 & 90.452 & 96.735 & 96.897 \\
\rowcolor{berry2}
\textbf{en} & 96.496 & 97.186 & 96.431 & 97.082 \\
\textbf{es} & 97.841 & 98.459 & 98.828 & 98.781 \\
\rowcolor{grey}
\textbf{et} & 95.142 & 95.244 & 96.547 & 97.402 \\
\textbf{eu} & 91.313 & 90.117 & 94.529 & 94.956 \\
\rowcolor{grey}
\textbf{fa} & 94.477 & 94.113 & 97.193 & 97.609 \\
\textbf{fi} & 93.534 & 93.436 & 96.275 & 97.823 \\
\rowcolor{grey}
\textbf{fr} & 96.970 & 97.112 & 97.845 & 98.102 \\
\textbf{fy} & 92.383 & 92.770 & 95.557 & 95.721 \\
\rowcolor{grey}
\textbf{ga} & 91.263 & 91.164 & 93.293 & 94.334 \\
\textbf{gd} & 91.774 & 90.220 & 92.814 & 93.797 \\
\rowcolor{grey}
\textbf{gl} & 94.261 & 95.565 & 95.130 & 96.892 \\
\textbf{he} & 91.373 & 90.847 & 96.242 & 97.035 \\
\rowcolor{grey}
\textbf{hi} & 83.566 & 94.437 & 96.554 & 97.384 \\
\textbf{hr} & 95.955 & 96.687 & 98.061 & 98.293 \\
\rowcolor{grey}
\textbf{hu} & 83.886 & 85.540 & 94.763 & 94.035 \\
\textbf{hy} & 53.953 & 86.965 & 92.985 & 93.829 \\
\rowcolor{grey}
\textbf{id} & 4.151 & 91.635 & 4.151 & 4.151 \\
\textbf{is} & 96.583 & 96.489 & 97.899 & 98.404 \\
\rowcolor{grey}
\textbf{it} & 96.521 & 97.297 & 98.172 & 98.334 \\
\textbf{ja} & 86.477 & 93.707 & 96.600 & 97.112 \\
\rowcolor{grey}
\textbf{ko} & 47.783 & 91.514 & 94.941 & 95.464 \\
\textbf{la} & 98.386 & 98.633 & 99.399 & 99.199 \\
\rowcolor{grey}
\textbf{lt} & 82.707 & 84.507 & 93.026 & 94.636 \\
\textbf{lv} & 93.252 & 93.710 & 95.744 & 96.908 \\
\rowcolor{grey}
\textbf{mt} & 87.284 & 85.603 & 89.248 & 86.349 \\
\textbf{nl} & 93.974 & 94.861 & 96.669 & 96.969 \\
\rowcolor{grey}
\textbf{pl} & 96.604 & 97.064 & 98.569 & 98.980 \\
\textbf{pt} & 95.420 & 96.572 & 97.526 & 97.611 \\
\rowcolor{grey}
\textbf{ro} & 95.795 & 96.044 & 97.568 & 97.878 \\
\textbf{ru} & 97.310 & 97.103 & 98.306 & 98.568 \\
\rowcolor{grey}
\textbf{sk} & 93.608 & 93.994 & 97.282 & 97.373 \\
\textbf{sl} & 95.233 & 95.776 & 98.157 & 98.798 \\
\rowcolor{grey}
\textbf{sr} & 96.140 & 97.154 & 98.531 & 98.802 \\
\textbf{sv} & 90.737 & 92.805 & 96.242 & 97.080 \\
\rowcolor{grey}
\textbf{ta} & 42.363 & 49.452 & 77.185 & 64.354 \\
\textbf{tr} & 92.390 & 92.320 & 94.667 & 94.946 \\
\rowcolor{grey}
\textbf{uk} & 93.320 & 93.980 & 96.020 & 96.975 \\
\textbf{ur} & 84.556 & 84.432 & 92.622 & 93.251 \\
\rowcolor{grey}
\textbf{vi} & 46.954 & 50.874 & 89.653 & 91.261 \\
\textbf{zh} & 55.811 & 84.877 & 95.022 & 95.972 \\
        \toprule
        \end{tabular}
    \caption{Full results for the finetuned POS tagging experiments in Section \ref{sec:cross-ling-exps}.}
    \label{tab:full-ft-pos}
\end{table*}